\documentclass[10pt,twocolumn,letterpaper]{article}

\usepackage[pagenumbers]{cvpr} 

\usepackage{graphicx}
\usepackage{amsmath}
\usepackage{amssymb}
\usepackage{booktabs}

\usepackage{multirow}
\usepackage{caption}
\usepackage{subcaption}
\usepackage[accsupp]{axessibility}

\usepackage[pagebackref,breaklinks,colorlinks]{hyperref}

\usepackage[capitalize]{cleveref}
\crefname{section}{Sec.}{Secs.}
\Crefname{section}{Section}{Sections}
\Crefname{table}{Table}{Tables}
\crefname{table}{Tab.}{Tabs.}

\def\cvprPaperID{1366} 
\def\confName{CVPR}
\def\confYear{2023}

\begin{document}

\title{Devil is in the Queries: Advancing Mask Transformers for Real-world Medical Image Segmentation and Out-of-Distribution Localization}
\author{
Mingze Yuan$^{1,2,\dagger}$, Yingda Xia$^{1, *}$, Hexin Dong$^{1,2}$, Zifan Chen$^{2}$, Jiawen Yao$^{1}$, Mingyan Qiu$^{1}$, Ke Yan$^{1}$, \\
Xiaoli Yin$^{4}$, Yu Shi$^{4}$, Xin Chen$^{3}$, Zaiyi Liu$^{3}$, Bin Dong$^{2,5}$, Jingren Zhou$^{1}$, Le Lu$^{1}$, Ling Zhang$^{1}$, Li Zhang$^{2}$ \\
$^{1}$Alibaba Group \,\,\,\,
$^{2}$Peking University \,\,\,\,
$^{3}$Guangdong Province People's Hospital \\
$^{4}$Shengjing Hospital\,\,\,\,
$^{5}$Peking University Changsha Institute for Computing and Digital Economy 
}

\maketitle

\begin{abstract}
Real-world medical image segmentation has tremendous long-tailed complexity of objects, among which tail conditions correlate with relatively rare diseases and are clinically significant. A trustworthy medical AI algorithm should demonstrate its effectiveness on tail conditions to avoid clinically dangerous damage in these out-of-distribution (OOD) cases. In this paper, we adopt the concept of object queries in Mask Transformers to formulate semantic segmentation as a soft cluster assignment. The queries fit the feature-level cluster centers of inliers during training. Therefore, when performing inference on a medical image in real-world scenarios, the similarity between pixels and the queries detects and localizes OOD regions. We term this OOD localization as MaxQuery. Furthermore, the foregrounds of real-world medical images, whether OOD objects or inliers, are lesions. The difference between them is less than that between the foreground and background, possibly misleading the object queries to focus redundantly on the background. Thus, we propose a query-distribution (QD) loss to enforce clear boundaries between segmentation targets and other regions at the query level, improving the inlier segmentation and OOD indication. Our proposed framework is tested on two real-world segmentation tasks, i.e., segmentation of pancreatic and liver tumors, outperforming previous state-of-the-art algorithms by an average of 7.39\% on AUROC, 14.69\% on AUPR, and 13.79\% on FPR95 for OOD localization. On the other hand, our framework improves the performance of inlier segmentation by an average of 5.27\% DSC when compared with the leading baseline nnUNet.
\let\thefootnote\relax\footnote{\text{*} Corresponding author. (yingda.xia@alibaba-inc.com)

$\quad \dagger$ Work was done during an internship at Alibaba DAMO Academy}
\end{abstract}
\vspace{-5mm}
\section{Introduction}
\label{sec:intro}

\begin{figure}[t]
  \centering
   \includegraphics[width=\linewidth]{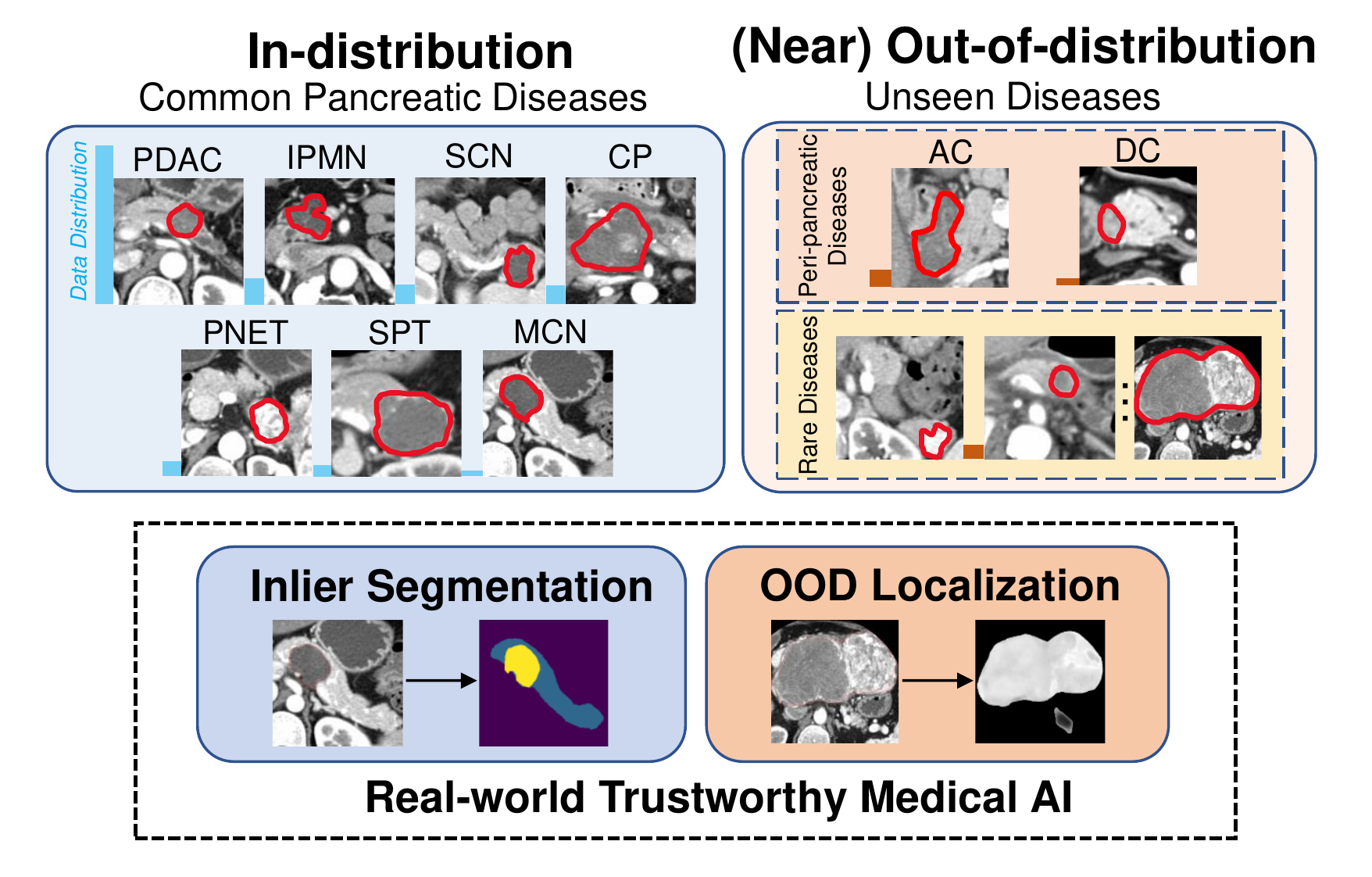}
   \caption{Real-world medical image segmentation. Real-world medical outliers (unseen, usually rare, tumors) are ``near" to the inliers (labeled lesions), forming a typical near-OOD problem. A real-world medical OOD detection/localization model should focus more on subtle differences between outliers and inliers than the significant difference between foreground and background..}
   \label{fig:task} 
   \vspace{-5mm}
\end{figure}

Image segmentation is a fundamental task in medical image analysis. With the recent advancements in computer vision and deep learning, automated medical image segmentation has reached expert-level performance in various applications ~\cite{kickingereder2019automated,yao2022deep,bian2022artificial}. Most medical image segmentation methods are based on supervised machine learning that heavily relies on collecting and annotating training data. However, real-world medical images are long-tailed distributed. The tail conditions are outliers and inadequate (or even unable) to train a reliable model~\cite{liu2020deep,zhao20213d,zhou2021review}. Yet, the model trained with inliers is risky for triggering failures or errors in real-world clinical deployment~ \cite{roy2022does}. For example, in pancreatic tumor image analysis, a miss-detection of metastatic cancer will directly threaten life; an erroneous recognition of a benign cyst as malignant will lead to unnecessary follow-up tests and patient anxiety. Medical image segmentation models should thus demonstrate the ability to detect and localize out-of-distribution (OOD) conditions, especially in some safety-critical clinical applications.

Previous studies have made valuable attempts on medical OOD localization~ \cite{tian2021constrained,zimmerer2022mood}, including finding lesions apart from normal cases or simulating OOD conditions for model validation. However, the real-world clinical scenario, such as tumor segmentation, is more complex, where either in-distribution or OOD cases have multiple types of tumors. Establishing a direct relationship between image pixels and excessive semantics (types of tumors) is difficult for real-world medical image segmentation. Using this relationship to distinguish inliers and outliers is even more challenging. Fortunately, several works about Mask Transformers~\cite{carion2020end,cheng2021per} have inspired us to split segmentation as a two-stage process of per-pixel cluster assignment and cluster classification~\cite{ yu2022cmt,yu2022k}. A well-defined set of inlier clusters may greatly benefit in identifying the OOD conditions from the medical images. Therefore, we propose MaxQuery, a medical image semantic segmentation framework that advances Mask Transformers to localize OOD targets. The framework adopts learnable object queries to iteratively fit inlier cluster centers. Since the affinity between OODs and an inlier cluster center should be less than that within the cluster (between inliers and cluster centers), MaxQuery uses the negative of such affinity as an indicator to detect OODs. 

Several recent works further define real-world medical image OOD localization as a near-OOD problem~\cite{winkens2020contrastive,roy2022does}, where the distribution gaps between inlier and OOD tumors are overly subtle, as shown in \cref{fig:task}. Thus, the near-OOD problems are more difficult. Our pilot experiments show that the cluster centers redundantly represent the large regions of background and organ rather than tumors, compromising the necessary variability of the cluster assignments for OOD localization. To solve this issue, we propose the query-distribution (QD) loss to regularize specific quantities of object queries on background, organ, and tumors. This enforces the diversity of the cluster assignments, benefiting the segmentation and recognition of OOD tumors.

We curate two real-world medical image datasets of (pancreatic and liver) tumor images from 1,088 patients for image segmentation and OOD localization. Specifically, we collect consecutive patients' contrast-enhanced 3D CT imaging with a full spectrum of tumor types confirmed by pathology. In these scenarios, the OOD targets are rare tumors and diseases. Our method shows robust performance across two datasets, significantly outperforming the previous leading OOD localization methods by an average of 7.39\% in AUROC, 14.69\% in AUPR, 13.79\% in FPR95 for localization, and 3.42\% for case-level detection. Meanwhile, our framework also improves the performance of inlier segmentation by an average of 5.27\% compared with the strong baseline nnUNet~\cite{isensee2021nnu}.

We summarize our main contributions as follows:
\begin{itemize}
    \item To the best of our knowledge, we are the first to explore the near-OOD detection and localization problem in medical image segmentation. The proposed method has a strong potential for utility in clinical practice.
    \item We propose a novel approach, MaxQuery, using the maximum score of query response as a major indicator for OOD localization. 
    \item A query-distribution (QD) loss is proposed to concentrate the queries on important foreground regions, demonstrating superior effectiveness for near-OOD problems.
    \item We curate two medical image datasets for tumor semantic segmentation/detection of real-world OODs. Our proposed framework substantially outperforms previous leading OOD localization methods and improves upon the inlier segmentation performance.
\end{itemize}
\section{Related Work}
\label{sec:related}

{\bf Medical Image Segmentation and Diagnosis.} U-Net~\cite{ronneberger2015u_unet0} and its variants~\cite{li2017h_unet1,liu20173d_unet2,milletari2016v_unet3,yu2017volumetric_unet4,zhou2019unet++_unet7} have been promoting the development of medical image segmentation. A recent self-configuring U-Net (nnUNet)~\cite{isensee2018nnu,isensee2021nnu} further surpassed existing approaches in various medical image segmentation tasks with minimal manual parameter tuning. Semantic segmentation serves as the core for downstream clinical tasks of disease detection~\cite{chu2019application}, differential diagnosis~\cite{de2018clinically,zhao20213d}, survival prediction~\cite{yao2022deep}, therapy planning~\cite{tang2019clinically}, and treatment response assessment~\cite{kickingereder2019automated}. Therefore, developing a reliable segmentation method is critical to improving safety in real-world clinical use. After the publication of Vision Transformers (ViTs)~\cite{dosovitskiy2020image}, integrating subsequent transformer blocks into the backbone of network architecture~\cite{chen2021transunet, hatamizadeh2022unetr,hatamizadeh2022swin,tang2022self_swinunetr} has been investigated. ViTs achieved improved results over traditional U-Net, particularly for multi-class semantic segmentation tasks. This work greatly focuses on exploring the real-world OOD localization detection problem over medical image segmentation. Current solutions provide limited performance, so we study a novel architecture combining Transformer and nnUNet for improving segmentation performance under clinical tasks, utilizing segmentation to detect and diagnose minority tumors \cite{zhao20213d}. 

{\bf Mask Transformers.} Unlike using Transformers directly as network backbones for natural and medical image segmentation~\cite{zheng2021rethinking,liu2021swin,xie2021segformer,strudel2021segmenter,yu2021glance}, Mask Transformers seek to enhance the CNN-based backbone with stand-alone transformer blocks. MaX-Deeplab~\cite{wang2021max} interprets object queries in DETR~\cite{carion2020end} as memory-encoded queries for end-to-end panoptic segmentation. MaskFormer~\cite{cheng2021per} further applies this design to semantic segmentation by unifying the CNN and the transformer branches. Afterward, Mask2Former~\cite{cheng2022masked} technically improves over its predecessor. Recently, CMT-Deeplab~\cite{yu2022cmt} and KMaX-Deeplab~\cite{yu2022k} propose to interpret the queries as clustering centers and add regulatory constraints for learning the cluster representations of the queries.  The design of Mask Transformers is intuitively suitable for medical image segmentation, especially for the semantic segmentation and diagnosis of tumors. This task requires the network to be locally sensitive to image textures for tumor segmentation and can globally understand organ-tumor morphological information for tumor sub-type recognition. To our knowledge, we are the first to adapt Mask Transformers for medical image segmentation and further explore its usage of recognizing outliers via queries. 

{\bf OOD Detection and Localization.} OOD Detection aims to detect the out-of-distribution conditions (outliers) that are unseen in the training data. Maximal softmax probability (MSP)~\cite{hendrycks2016baseline} serves as a strong baseline. After that, various approaches improved OOD detection from multiple aspects~\cite{ood2,ood3,ood4,ood5}. These approaches focus on image-level OOD detection, and efforts have also been made to localize OOD objects or regions on a large image, e.g., urban driving scenes~\cite{hendrycks2016baseline,blum2019fishyscapes, lis2019detecting, chan2021segmentmeifyoucan, xia2020synthesize, jung2021standardized,raml2022,oberdiek2020detection}. Despite the advance of OOD detection and localization on natural images, its application on real-world medical images is challenging. Since the difference between foregrounds in real-world medical images is subtle, their OOD detection/localization becomes a typical near-OOD problem~\cite{winkens2020contrastive,ren2021simple,mirzaei2022fake,dong2022neural}. Therefore, the existing OOD solutions could hardly be recommended for clinical practice~\cite{tian2021constrained,zimmerer2022mood,pinaya2022unsupervised}. Recent work, HOD~\cite{roy2022does}, paces one step toward real-world OOD detection of rare diseases in dermatology classification.

\section{Method}
\label{sec:method}
In this section, we first provide an overview of our method and then describe our proposed query-distribution (QD) loss and MaxQuery framework for OOD localization. 
\subsection{Method Overview}
Medical image segmentation aims to segment an image into multiple regions representing anatomical objects of interest. Here, we focus on 3D medical image $\mathbf{X} \in \mathbb{R}^{H \times W \times D}$, and use a segmentation model to partition it into $K$ category-labeled binary masks,
\begin{equation}
\label{eq:ground-truth}
    \mathbf{G} = \{\mathbf{G}_{i}\}_{i=1}^{K},
\end{equation}
where $\mathbf{G}_{i} \in \{0, 1\}^{H \times W \times D}$ is the ground truth mask that belongs to the $i$-th class, and $\sum_{i=1}^{K} \mathbf{G}_{i} = \mathbf{1}^{H \times W \times D}$. In our problem, class 1 refers to background, class 2 stands for specific organ, and the others for tumors. Since the real-world medical image dataset has a long-tail distribution in quantity, its segmentation task should be divided into supervised inlier segmentation and pixel-level OOD localization.
\begin{figure*}[t]
  \centering
  \includegraphics[width=0.9\linewidth]{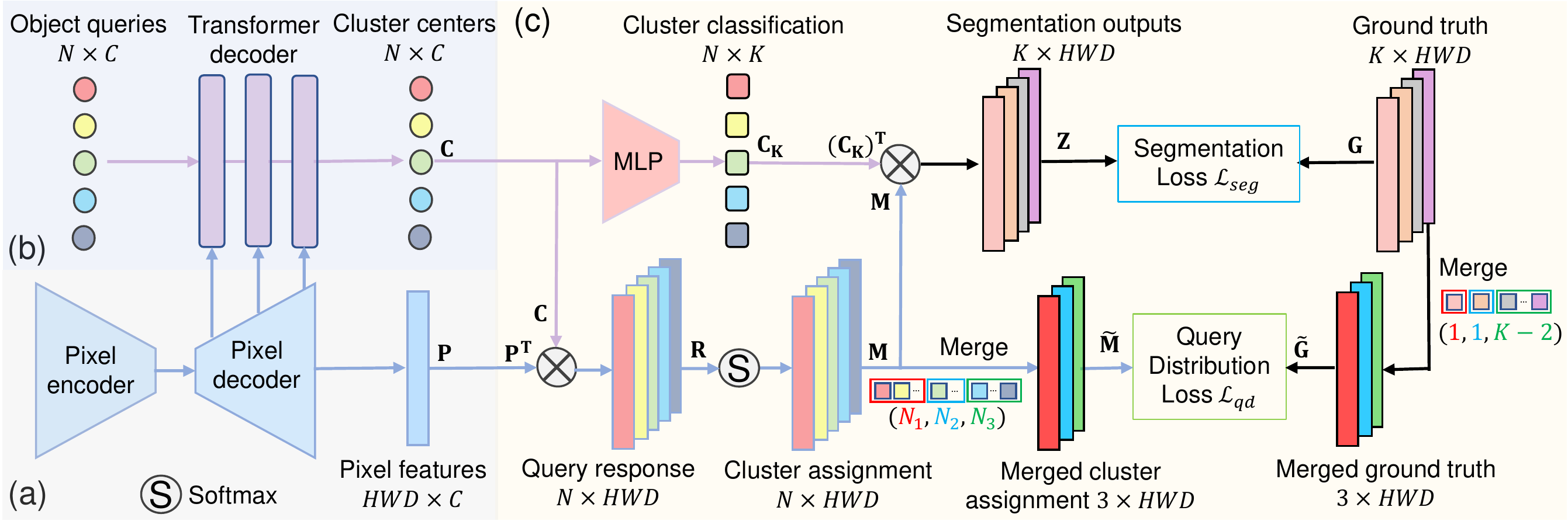}
  \caption{Overview of our proposed framework. (a) A CNN backbone for image segmentation, here we use nnUNet~\cite{isensee2021nnu}; (b) A transformer decoder interatively updates the object queries to fit the inlier cluster centers; (c) A two-stage cluster analysis: 1) cluster assignment groups the pixels based on the affinity between pixel features and cluster centers; 2) cluster classification guides the grouped pixels to generate segmentation logits. The overall segmentation is supervised by a classic segmentation loss and a novel query-distribution loss.}
  \label{fig:method}
  \vspace{-3mm}
\end{figure*}

{\bf Inlier Segmentation.} As shown in Fig.~\ref{fig:method}, we build our model with a CNN backbone to extract per-pixel features $\mathbf{P} \in \mathbb{R}^{HWD \times C}$ and a transformer module. The transformer module gradually updates a set of learnable object queries, $\mathbf{C} \in \mathbb{R}^{N \times C}$, to meaningful mask embedding vectors via cross attention between object queries and per-pixel features, 
\begin{equation}
    \mathbf{C} \gets \mathbf{C} +  \operatornamewithlimits{argmax}_{N} (\mathbf{Q}^c (\mathbf{K}^{p})^{\mathrm{T}}) \mathbf{V}^{p},
\end{equation}
where the superscripts $c$ and $p$ represent query and pixel features, respectively. We also adopt cluster-wise argmax from KMax-DeepLab~\cite{yu2022k} to substitute spatial-wise softmax in the original cross attention settings.

Inspired by recent works on cluster analysis of mask transformers~\cite{yu2022cmt,yu2022k}, we consider semantic segmentation as a two-stage cluster analysis process. First, all pixels are assigned into different clusters. The mask embedding vectors $\mathbf{C}$ from the transformer module are formulated as the cluster centers. The product $\mathbf{R}$ of $\mathbf{C}$ and $\mathbf{P}^{\mathrm{T}}$ represents the query response, which expresses the similarity between each pixel and cluster centers. Then, we use the query-wise softmax activation on query responses $\mathbf{R}$ to generate a mask prediction, which encourage the exclusiveness of cluster assignment. The mask prediction (cluster assignment) $\mathbf{M}$ is defined as,
\begin{equation}
\label{eq:query-response}
    \mathbf{M} = \operatornamewithlimits{softmax}_{N}(\mathbf{R}) =\operatornamewithlimits{softmax}_{N}(\mathbf{C} \mathbf{P}^{\mathrm{T}}).
\end{equation} 
Notably, different from the sigmoid activation used in~\cite{cheng2021per,cheng2022masked}, the query-wise softmax activation could better guide the object queries (cluster centers) to focus on different regions of the image and encourage diversity in real-world medical image segmentation.

Secondly, the grouped pixels are classified under the guidance of cluster classification. We evaluate the cluster centers $\mathbf{C}$ via a multi-layer perceptron (MLP) to predict the $K$-channel cluster classifications $\mathbf{C}_{K} \in \mathbb{R}^{N \times K}$ for all $N$ clusters. We then aggregate the cluster assignments $\mathbf{M}$ of grouped pixels and their classifications $\mathbf{C}_{K}$ for the final semantic segmentation,
\begin{equation}
\label{eq:final_seg}
    \mathbf{Z} = (\mathbf{C}_{K})^{\mathrm{T}} \mathbf{M},
\end{equation}
where $\mathbf{Z} \in \mathbb{R}^{K \times HWD}$ represents the final logits. To supervise the final segmentation, we combine the classic segmentation loss and a novel QD loss between final output $\mathbf{Z}$ and ground truth $\mathbf{G}$, more details in Sec~\ref{sec:qd_loss}.

{\bf OOD Localization.} To further segment abnormal regions unseen in training images, an OOD localization process is required when performing inference on a test image. Formally, given a test image $\mathbf{X} \in \mathbb{R}^{H \times W \times D}$, OOD localization evaluates the query response to find the maximal one that represents the similarity between the pixel and its assigned cluster center. Then, the model can generate an pixel-wise anomalous score map $\mathbf{A} \in [0, 1]^{H \times W \times D}$, where $\mathbf{A}_{i} = 1$ and $\mathbf{A}_{i} = 0$ represent that $i$-th pixel in $\mathbf{X}$ belongs to an OOD class and an in-distribution class, respectively. More details of this novel OOD localization (MaxQuery) is in Sec~\ref{sec:method_ood}

\subsection{Managing Cluster Distribution with QD Loss}
\label{sec:qd_loss}
Classic segmentation loss serves as an important learning target of our model. We combine the Cross-Entropy and Dice losses between final output $\mathbf{Z}$ and ground truth $\mathbf{G}$ in \cref{eq:ground-truth} as the  segmentation loss, \textit{i.e.}, $\mathcal{L}_{seg} = \ell_{ce} + \ell_{dc}.$ However, when only using classic segmentation loss, object queries focus majorly on the background and organs rather than the tumors. The significant difference between foreground and background greatly distracts the model from focusing on subtle differences between OOD objects and inliers. As later shown in an example in Fig~\ref{fig:vis_wo_qd} , some queries may even have mixed representation on background and foreground which is an unsatisfactory phenomenon for discriminative cluster learning. 
Therefore, we propose query-distribution (QD) loss to manipulate the object queries and guide them to focus on the foreground, especially the tumors, and encourage concentrated cluster learning. The key idea is to use ground-truth $\mathbf{G} \in \mathbb{R}^{K \times HWD}$ to supervise the cluster assignment probability maps $\mathbf{M}$. This motivation also benefits OOD localization as introduced in Sec~\ref{sec:method_ood}.

We thus divide the $N$ channels into three groups, including $N_{1}, N_{2}, N_{3}$ queries, for background, organ and tumor regions, respectively.  
Our goal is to associate the first $N_1$ channels of $\mathbf{M}$ (representing the assignment probabilities of the first $N_1$ cluster centers) with the background class $\mathbf{G}_1$, the next $N_2$ channels with the organ class $\mathbf{G}_2$, and the last $N_3$ channels with the tumor classes $\sum_{i=3}^{K}\mathbf{G}_i$. We define the merged cluster assignments $\Tilde{\mathbf{M}}$ and class labels $\Tilde{\mathbf{G}}$ as the following,
\begin{equation}
\begin{split}
    \Tilde{\mathbf{M}} &= (\Tilde{\mathbf{M}}_1,\Tilde{\mathbf{M}}_2,\Tilde{\mathbf{M}}_3)  \in \mathbb{R}^{3 \times HWD}\\
   &=(\sum_{i=1}^{N_1}\mathbf{M}_i, \sum_{j=1}^{N_{2}}\mathbf{M}_{N_{1}+j},
   \sum_{k=1}^{N_{3}}\mathbf{M}_{N_{1}+N_{2}+k}),
\end{split}
\end{equation}
\begin{equation}
    \Tilde{\mathbf{G}} = (\Tilde{\mathbf{G}}_1,\Tilde{\mathbf{G}}_2,\Tilde{\mathbf{G}}_3) = (\mathbf{G}_1, \mathbf{G}_2,
   \sum_{i=3}^{K}\mathbf{G}_i)  \in \mathbb{R}^{3 \times HWD},
\end{equation}
where the merged $\Tilde{\mathbf{M}}$ are still probability distributions in each spatial position, \textit{i.e.}, $\sum_{j=1}^{3} \Tilde{\mathbf{M}}_{i} = \mathbf{1}^{H\times W \times D}$. 

Finally, we formulate the QD loss as the negative log likelihood loss between $\Tilde{\mathbf{M}}$ and $\Tilde{\mathbf{G}}$,
\begin{equation}
    \mathcal{L}_{qd} = -\sum_{j=1}^{HWD} \sum_{i=1}^{3} \Tilde{\mathbf{G}}_{ij}\log \Tilde{\mathbf{M}}_{ij},
\end{equation}
which draws strict boundaries between different types of cluster assignments ($\Tilde{\mathbf{M}}_1$, $\Tilde{\mathbf{M}}_2$, and $\Tilde{\mathbf{M}}_3$) based on the ground truth.
The final loss function $\mathcal{L}$ is a combination of segmentation loss $\mathcal{L}_{seg}$ and QD loss $\mathcal{L}_{qd}$ with a balance weight $\lambda$, formulated as,
\begin{equation}
    \mathcal{L} = \mathcal{L}_{seg} + \lambda \mathcal{L}_{qd}.
\end{equation}

\subsection{Localizing OOD Regions with MaxQuery}
\label{sec:method_ood}

\begin{figure}[t]
  \centering
   \includegraphics[width=0.7\linewidth]{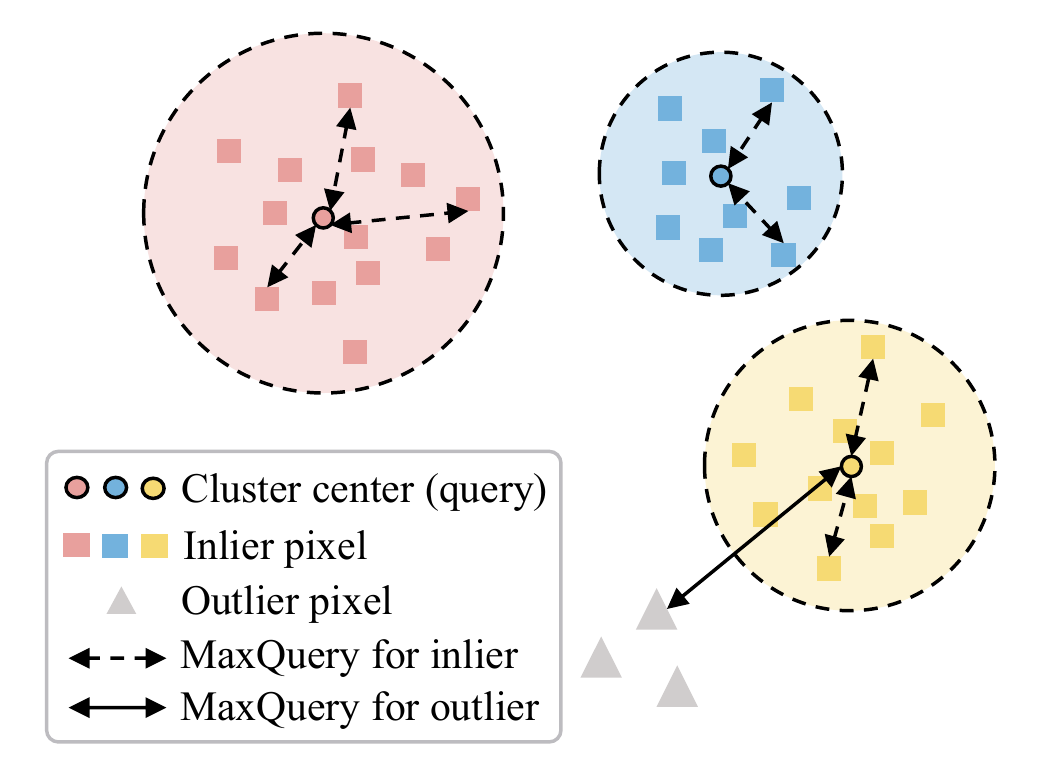}
   \vspace{-2mm}
   \caption{Illustration of how MaxQuery works. MaxQuery, \textit{i.e.}, the negative of maximal query response, reflects the distance of the pixel and its assigned cluster center. MaxQuery of the inlier (dotted arrow) is usually smaller than that of the outlier (solid arrow) and thus is able to identify the anomalous/OOD pixels.}
   \label{fig:maxquery}
   \vspace{-3mm}
\end{figure} 

Given a test image $\mathbf{X} \in \mathbb{R}^{H \times W \times D}$, our mask transformer will yield the pixel-level query response $\mathbf{R} \in \mathbb{R}^{N \times H \times W \times D}$, representing the affinity of pixel feature and cluster centers. The maximal query response of one pixel then represents the similarity between the pixel and its assigned cluster center. Intuitively, the maximal query response of outliers should be smaller than inliers. We therefore adopt the negative of maximal query response in \cref{eq:query-response} as the pixel-wise anomaly score, called MaxQuery, \textit{i.e.},
\begin{equation}
    \label{eq:anom-score}
    \mathbf{A} = -\max_{N} \mathbf{R},
\end{equation}
where $\mathbf{R} \in \mathbb{R}^{N \times H \times W \times D}$ represents the query response matrix and $\mathbf{A} \in \mathbb{R}^{H \times W \times D}$ indicates the anomaly score map. The anomaly score can be further normalized into $[0, 1]$ by min-max normalization.  \Cref{fig:maxquery} illustrates the capability of MaxQuery for OOD pixels identification. The subscript $N$ means that we perform maximum operation on the query dimension. We add a minus sign because when the maximal query response of a pixel is larger, it is less likely to be an OOD pixel.

In addition, we compare the results of anomaly score maps according to the maximum of query responses $\mathbf{R}$ (pre-softmax, $\mathbf{A}=-\max_{N} \mathbf{R}$) and cluster assignments $\mathbf{M}$ (post-softmax, $\mathbf{A}'=-\max_{N} \mathbf{M}$). $\mathbf{A}$ greatly outperforms $\mathbf{A}'$, since if an inlier pixel is evenly close to multiple cluster centers, the maximal score in $\mathbf{M}$ can be very low and easily be mis-classified as an outlier. But with maximum query response $\mathbf{R}$ (pre-softmax), the score is still high enough for an indication of inlier. Thus we choose the maximal query response to imply the anomalous regions.

\vspace{-0.2cm}

\section{Experiments}
\label{sec:experiments}
\subsection{Datasets and Experiment Setting}
We collect two datasets, i.e., pancreas and liver tumor segmentation datasets, which includes contrast-enhanced 3D CT scans from consecutive patients before treatment.  We register the multi-phase CT scans into arterial late and venous phase using DEEDS~\cite{heinrich2013mrf}, respectively. All types of tumors are confirmed by pathology, except for cysts in the liver (confirmed by a radiologist specialized in liver imaging). All tumors are annotated slice-by-slice manually on the CT phase with the best tumor visibility by experienced radiologists specialized in specific diseases. The organ (pancreas or liver) in each dataset is first annotated automatically by a self-learning approach \cite{zhang2018self} trained on public datasets (e.g., Medical Decathlon \cite{antonelli2022medical}) and then edited by engineers.

\begin{table*}
\centering
\small
\resizebox{\textwidth}{!}{
\begin{tabular}{l|cccc|cccc}
\hline
                         & \multicolumn{4}{c|}{Pancreatic \%}                                                       & \multicolumn{4}{c}{Liver \%}                                                             \\ \hline
\multirow{2}{*}{Methods} & \multicolumn{3}{c|}{OOD Localization}     & \multicolumn{1}{c|}{$\mathrm{OOD}_{\mathrm{case}}$}    & \multicolumn{3}{c|}{OOD Localization}     & \multicolumn{1}{c}{$\mathrm{OOD}_{\mathrm{case}}$}     \\ \cline{2-9} 
                         & AUROC$\uparrow$ & AUPR$\uparrow$ & \multicolumn{1}{c|}{$\mathrm{FPR}_{\mathrm{95}}\downarrow$} & \multicolumn{1}{c|}{AUC$\uparrow$}       & AUROC$\uparrow$ & AUPR$\uparrow$ & \multicolumn{1}{c|}{$\mathrm{FPR}_{\mathrm{95}}\downarrow$} & \multicolumn{1}{c}{AUC$\uparrow$} \\ \hline
MC Dropout~\cite{kendall2017uncertainties}          & 49.08     &11.47      & \multicolumn{1}{c|}{84.60}      & \multicolumn{1}{c|}{72.91}                  &   39.61    & 16.05     & \multicolumn{1}{c|}{91.13}      & \multicolumn{1}{c}{34.05}                \\
MSP~\cite{hendrycks2016baseline}                      &  53.81     &  13.44    & \multicolumn{1}{c|}{86.44}      & \multicolumn{1}{c|}{73.38}                  &  75.14     & 25.27     & \multicolumn{1}{c|}{70.04}      & \multicolumn{1}{c}{66.76}              \\
MaxLogit ~\cite{hendrycks2019scaling}                &  58.46     &21.93      & \multicolumn{1}{c|}{83.68}      & \multicolumn{1}{c|}{73.42}                & 78.60      &    35.47  & \multicolumn{1}{c|}{48.73}      & \multicolumn{1}{c}{65.68}                 \\
SynthCP ~\cite{xia2020synthesize}                 &  69.86     &    26.50  & \multicolumn{1}{c|}{66.65}      & \multicolumn{1}{c|}{68.43}                 &74.93       &  34.03    & \multicolumn{1}{c|}{57.91}      & \multicolumn{1}{c}{63.34}               \\
SML ~\cite{jung2021standardized}                     & 56.10      & 30.44     & \multicolumn{1}{c|}{77.81}      & \multicolumn{1}{c|}{62.26}                & 86.64      &44.59      & \multicolumn{1}{c|}{31.04}      & \multicolumn{1}{c}{63.85}               \\ \hline
Ours (w/o $\mathcal{L}_{qd}$)                    &  63.54     & 25.25     & \multicolumn{1}{c|}{67.09}      & \multicolumn{1}{c|}{74.87}                 & 74.95      &42.31      & \multicolumn{1}{c|}{53.52}      & \multicolumn{1}{c}{65.91}                  \\
\textbf{Ours}                     &\textbf{82.52}       &\textbf{55.60 }     & \multicolumn{1}{c|}{\textbf{46.19}}      & \multicolumn{1}{c|}{\textbf{77.97}}                   &   \textbf{88.75}    & \textbf{48.80}     & \multicolumn{1}{c|}{\textbf{23.93}}      & \multicolumn{1}{c}{\textbf{69.04}}           \\
\hline
\end{tabular}
}
\caption{OOD localization and case-level OOD detection performance on \textit{Pancreatic Tumors} and \textit{Liver Tumors}. Our proposed method achieves state-of-the-art OOD detection performance at both pixel level and case level. All the methods are implemented based on the nnUNet~\cite{isensee2021nnu} backbone. ($\mathrm{OOD}_{\mathrm{case}}$: case-level OOD detection.)}
\label{tab:main}
\end{table*}

\begin{table*}[ht]
\centering
\resizebox{\textwidth}{!}{
\begin{tabular}{l|cccccccc|cccccc}
\hline
           & \multicolumn{8}{c|}{Pancreatic \%}                                         & \multicolumn{6}{c}{Liver \%}                                                      \\ \hline
Methods    & PDAC & IPMN & PNET & SCN & CP & SPT & \multicolumn{1}{c|}{MCN} & Avg. & HCC & ICC & Meta. & Heman. & \multicolumn{1}{c|}{Cyst} & Avg.                  \\ \hline
nnUNet~\cite{isensee2021nnu}   & 65.65     &  27.60    &32.59      & 36.46    &  23.33  & 31.73  &     \multicolumn{1}{c|}{30.96}    & 35.47     & 57.22    &  28.16   & 52.81      &   77.55     & \multicolumn{1}{c|}{46.49}     &          52.45             \\
Ours (w/o $\mathcal{L}_{qd}$)       &  65.87 & 28.3 & 32.43 & 40.63 & 28.93 & 30.77 &      \multicolumn{1}{c|}{30.89}    & 36.84    & 60.91   & 30.58    & 53.21     & 78.47      & \multicolumn{1}{c|}{46.42}     & 53.92\\
\textbf{Ours}       &    67.91     & 46.92     & 32.07     &    42.51 &31.36    &42.67     &      \multicolumn{1}{c|}{28.97}    &   \textbf{41.77}   &  67.61   &  30.78   &  60.40     &  77.07      & \multicolumn{1}{c|}{47.61}     & \textbf{56.69} \\ \hline
\end{tabular}
}
\caption{Inlier segmentation Dice scores (\%) on \textit{val} set of \textit{Pancreatic Tumors} and \textit{Liver Tumors} (all methods report results with final checkpoint). Compared with the benchmark model (nnUNet~\cite{isensee2021nnu}) in medical image segmentation, our method noticeably outperforms the strong baseline for the task of inlier tumor segmentation. See the Appendix for other baselines.}
\label{tab:seg} 
\vspace{-0.1cm}
\end{table*}

{\bf Pancreatic Multi-type Tumors} dataset contains 661 patients. Every patient has five phases of CT scans: noncontrast, arterial-early, arterial-late, venous, and delay. The median spacing is $3 \times 0.419 \times 0.419$ mm. According to previous clinical studies about pancreatic tumor classification \cite{springer2019multimodality,chu2022classification}, we assign the seven most common conditions (PDAC, PNET, SPT, IPMN, MCN, CP, and SCN) as inliers, and allocate AC, DC, and ``other'' as outliers. We randomly split 590 inlier data into 378(64\%) training, 94(16\%) validation, and 118(20\%) testing, and leave out all 71 outlier data for OOD testing.

{\bf Liver Multi-type Tumors } dataset contains 427 patients. Each patient has three phases of CT scans: noncontrast, arterial, and venous. The median spacing is $3 \times 0.760 \times 0.760$ mm. Following \cite{yasaka2018deep}, we assign the five most common conditions (HCC, ICC, metastasis, hemangiomas, and cyst) as inliers, and allocate hepatoblastoma, FNH, and ''other'' as outliers. Similarly, We randomly split 327 inlier data into 209(64\%) training, 52(16\%) validation, and 66(20\%) testing, and leave out all 100 outlier data for OOD testing. Notice that the ``other'' class in both datasets contains multiple rare diseases, reflecting the long-tailed distribution of real-world disease incidence.

\subsection{Implementation \& Evaluation Metrics}

{\bf Network Architecture.} We use the current benchmark model in medical image segmentation, nnUNet~\cite{isensee2021nnu}, as a CNN backbone, which consists of a pixel encoder and a pixel decoder with skip connections. We adopt four transformer decoder blocks, and each takes pixel features with output stride 32, 16, 8, and 4, respectively. The self-attention layer in the block has 8 heads. Since medical image segmentation is sensitive to local textures, we add a decoder block for output stride 4 compared with previous works~\cite{wang2021max,yu2022k}. To increase numerical stability, we add an InstanceNorm~\cite{ulyanov2016instance} layer and a LayerNorm~\cite{ba2016layer} at the end of pixel-level and transformer decoder modules, respectively.

{\bf Training and Testing.} Each CT scan is resampled into the median spacing per tumor dataset (\textit{e.g.}, $3 \times 0.419 \times 0.419$ mm for the pancreatic dataset) and  normalized into zero mean and unit variance. Our model is trained using a batch size of 2 on one GPU (with $28 \times 192 \times 320$ patch size for pancreatic, $40 \times 192 \times 224$ for liver). We adopt the drop path~\cite{huang2016deep} strategy with a probability of 0.2 for regularization. During training, extensive data augmentation is utilized on-the-fly~\cite{isensee2021nnu} to improve the generalization, including random rotation and scaling, elastic deformation, additive brightness, and gamma scaling. The network is trained with RAdam~\cite{liu2019variance} with the initial learning rate as $1 \times 10^{-4}$ and a polynomial learning rate decay. We first pre-train the nnUNet backbone for 1000 epochs and finetune the whole architecture jointly for another 200 epochs. During finetuning, we keep the backbone weights fixed for the first 50 epochs, and then set it with a learning rate multiplier of 0.1 for the next 150 epochs. The number of object queries (\textit{i.e.}, cluster centers) $N$ is 32, and the query distribution $(N_1, N_2, N_3)$ is set as (16, 4, 12). We follow KMax-DeepLab~\cite{yu2022k} to directly add deep supervision on the attention map of ($k$-means) cross attention to align it with the final segmentation after the segmentation output head. The loss weight $\lambda$ for QD loss is 0.1. 

{\bf Evaluation Metrics.} For OOD localization, we follow the standard metrics for anomaly segmentation~\cite{xia2020synthesize,jung2021standardized,raml2022}. We compute the area under receptive-operative curve (AUROC) and the area under precision-recall curve (AUPR). We also report FPR at the TPR level of 0.95 (FPR95) as OOD localization metrics since the false positive rate is safety-critical in clinical practice. For case-level OOD detection, we compute the average of anomaly scores in predicted tumor regions as the case-level anomaly score and choose AUC as the case-level OOD detection metric. Meanwhile, we report the average Dice Score of inlier tumors to evaluate the segmentation performance on inlier classes.

{\bf Baselines.} For OOD localization, we compare our work with a series of representative anomaly segmentation methods in multiple aspects, including uncertainty statistics-based (MSP~\cite{hendrycks2016baseline}, MaxLogit~\cite{hendrycks2019scaling}, SML~\cite{jung2021standardized}), Bayesian deep learning-based (MC Dropout~\cite{kendall2017uncertainties}) and image re-synthesis-based (SynthCP~\cite{xia2020synthesize}) methods. All of them are implemented using nnUNet~\cite{isensee2021nnu} backbone. For inlier segmentation, we compare our work with the benchmark model (nnUNet~\cite{isensee2021nnu}) and previous leading model (Swin UNETR~\cite{tang2022self_swinunetr}), as well as UNet~\cite{ronneberger2015u_unet0}, UNet++~\cite{zhou2019unet++_unet7} and TransUNet~\cite{chen2021transunet}, implemented by their officially released code and pre-trained model with same settings.

\subsection{Main Results}
\label{sec:results}
\begin{figure*}[t]
  \centering
   \includegraphics[width=0.8\linewidth]{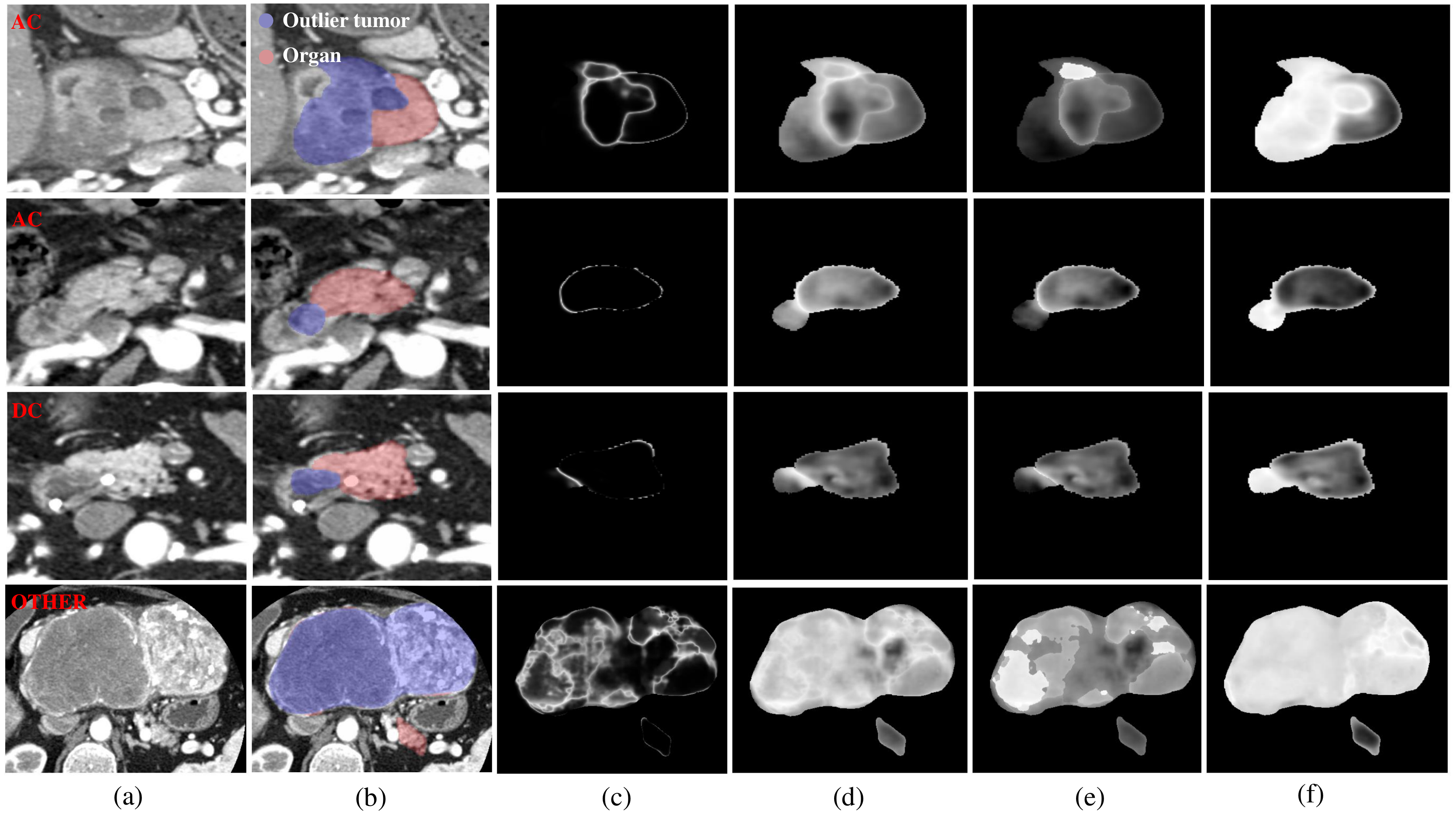}
   \caption{Visualization results of anomaly score map for OOD localization on \textit{Pancreatic Tumors}: (a) CT slice, (b) ground truth (red: pancreas, blue: outlier tumor), (c) MSP~\cite{hendrycks2016baseline}, (d) MaxLogit~\cite{hendrycks2019scaling}, (e) SML~\cite{jung2021standardized} and (f) Ours. The grayscale level indicates the anomaly score. Our approach maintains a high anomaly score in the OOD pixels (outlier tumor), while a low anomaly score in the in-distribution pixels (organ). The four cases are selected from three different unknown diseases to show our method's robustness to tumor type.}
   \label{fig:comparison}
\end{figure*}

\begin{figure}[ht]
  \centering
  
  \begin{subfigure}[a]{0.42\textwidth}
         \centering
         \includegraphics[width=\textwidth]{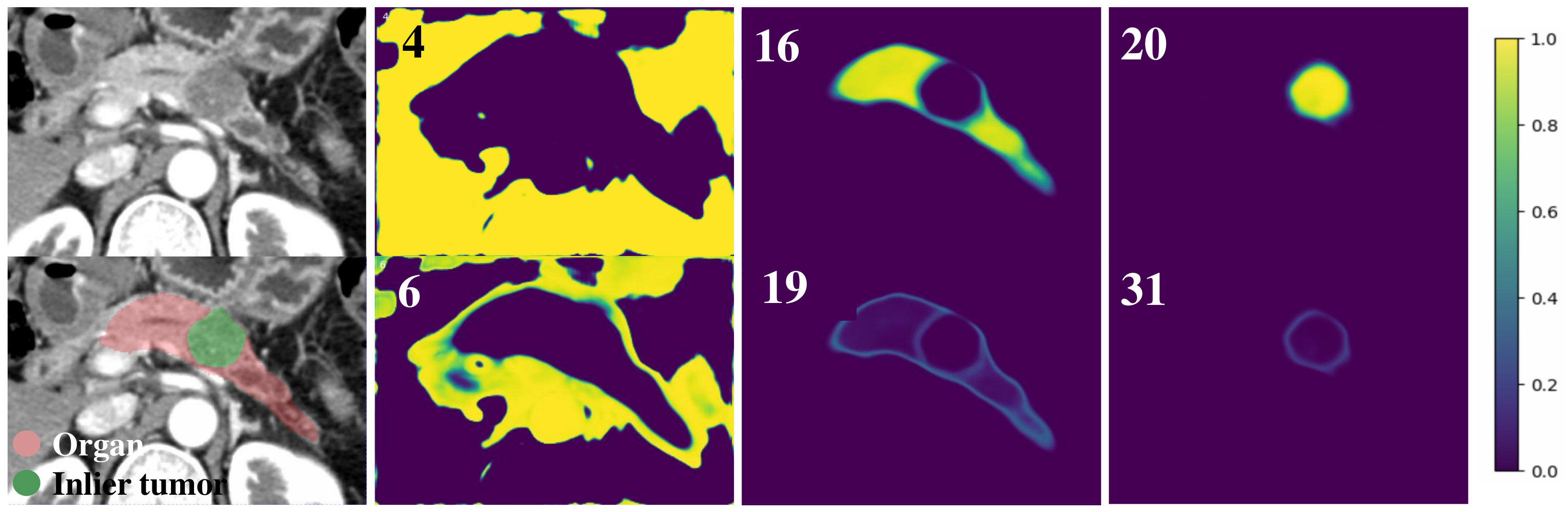}
         \caption{An in-distribution example.}
         \label{fig:in-lier}
     \end{subfigure}
     
\begin{subfigure}[b]{0.42\textwidth}
         \centering
         \includegraphics[width=\textwidth]{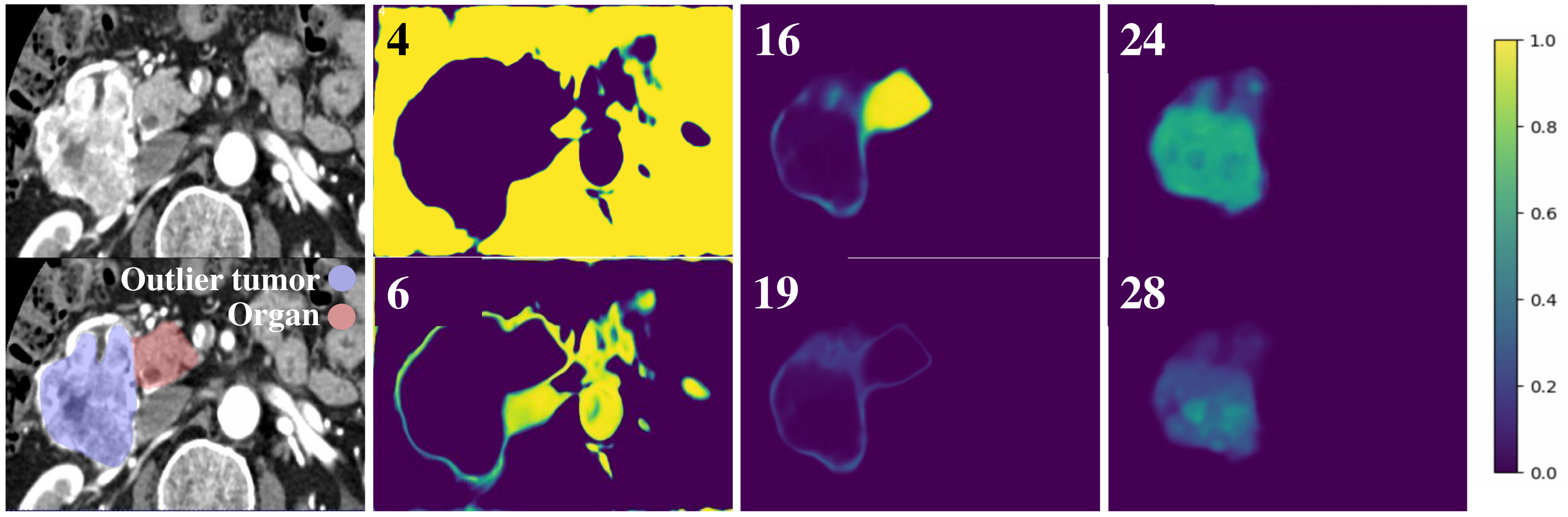}
         \caption{An out-of-distribution example.}
         \label{fig:outlier}
     \end{subfigure}
\caption{Visual examples of cluster assignments for (a) an in-distribution and (b) an out-of-distribution (OOD) sample. From left to right: (Column 1) image and ground truth with red: organ, green: inlier tumor, blue: outlier tumor; (Columns 2-4) representative object queries for background (C2), organ (C3) and tumor (C4), respectively. Query IDs are at the upper-left corners.}
  \label{fig:vis_mask} 
\end{figure}

Comparisons on the real-world datasets, including \textit{Pancreatic Tumors} and \textit{Liver Tumors}, are summarized in \cref{tab:main,tab:seg}. We also present visualization examples in \cref{fig:comparison,fig:vis_mask} to better understand the role of object queries in our proposed mask transformer and compare different anomaly segmentation methods. 

{\bf Pancreatic Tumors.} In \cref{tab:main}, we compare MaxQuery with other baselines on \textit{Pancreatic Tumors}. 
Our framework shows the best performance in all metrics. Specifically, our framework outperforms the previous best method SML~\cite{jung2021standardized} by a large margin of 12.66\% in AUROC, 25.16\% in AUPR, 20.46\% in FPR95 for OOD localization, and 4.55\% in AUC for case-level OOD detection. For qualitative analysis, we present four visual examples from \textit{Pancreatic Tumors} by visualizing the anomaly score map of MSP~\cite{hendrycks2016baseline}, MaxLogit~\cite{hendrycks2019scaling}, SML~\cite{jung2021standardized}, and ours. As shown in \cref{fig:comparison}, our method maintains a high anomaly score in the OOD pixels (outlier tumor), while a low anomaly score in the in-distribution pixels (organ). Moreover, the previous methods underestimate the anomalous score map. They tend only to highlight the boundaries of the OOD region, but our method preserves a high anomalous score on the entire OOD region. 

In \cref{tab:seg}, our segmentation performance for inliers surpasses nnUNet by 6.30\% in DSC. These improvements demonstrate that our framework can simultaneously detect common diseases with high accuracy and identify rare diseases in pixel-level localization and case-level diagnosis without requiring very large data samples. (Other baselines can be found in the Appendix.)

We visualize the mask predictions of in-distribution and OOD examples to illustrate the working mechanism of object queries as cluster centers and how MaxQuery identifies the OOD condition. As shown in \cref{fig:vis_mask}, for either in-distribution or OOD example, the background and organ regions are confidently activated by specific queries (Queries 4 and 6 for background, Query 16 for the target organ). Interestingly, regions with distinguishing features, such as the aorta or other abdominal organs, are not activated by the major cluster center (Query 4) but by an independent center (Query 6). This supports that the queries gradually converge to different meaningful centers. 
Furthermore, the corresponding queries of specific in-distribution tumors usually concentrate at a single center (Query 20 in \cref{fig:in-lier}). Yet, queries corresponding to the OOD tumors seem to split into multiple centers with lower responses (Query 24 and 28 in ~\cref{fig:outlier}). The visual examples fulfill the motivation of the proposed MaxQuery that no inlier cluster centers can dominantly fit the OOD pixels.

\textbf{Liver Tumors.} \Cref{tab:main} also shows the quantitative result on \textit{Liver Tumors}. Our method outperforms the baselines in all evaluation metrics. Note that SML~\cite{jung2021standardized} improves the performance in OOD localization while dropping its performance in case-level OOD detection compared with MaxLogit~\cite{hendrycks2019scaling}, whereas our method performs well in both pixel and case level. Particularly, our method reaches a significantly lower FPR95 of 23.93\% compared with previous approaches, which is crucial to localizing the OOD regions in medical scenarios. As shown in \cref{tab:seg}, our segmentation performance for inliers surpasses nnUNet by 4.24\% in DSC. The qualitative analysis on \textit{Liver Tumors} is in the Appendix.

\begin{figure}[ht]
  \centering
  \includegraphics[width=.8\linewidth]{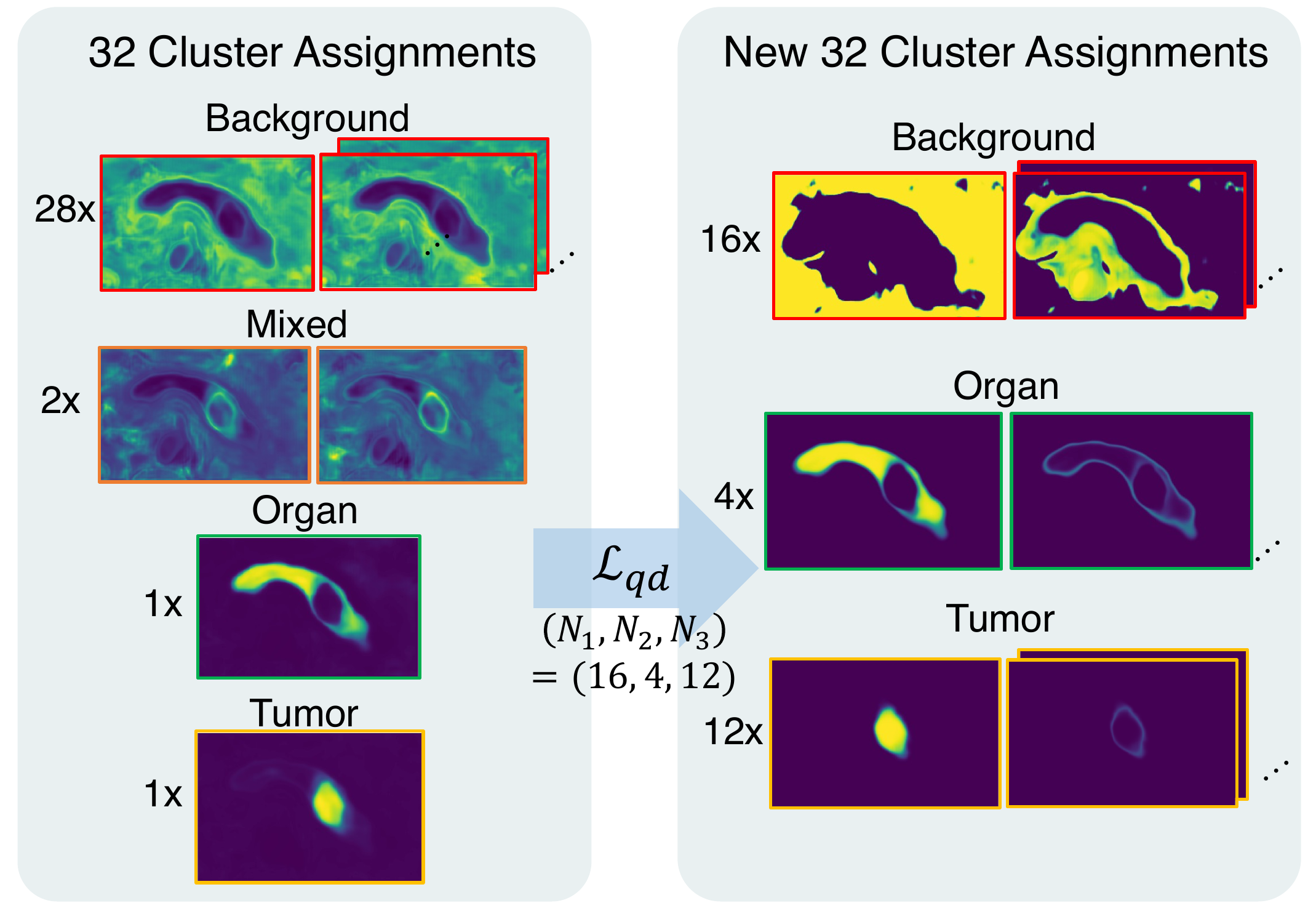}
  \caption{The effect of QD loss by visualizing the cluster assignment maps of the 32 queries on an inlier. {\bf Left:} without QD loss, most queries redundantly focus on the background and some queries mix the background with foreground. {\bf Right:} after using QD loss, we can manage the query distribution on the background, organ, and tumor with better seperation. The clear boundaries and high responses shows that QD loss encourages discriminative representation learning of the queries which will benefit both segmentation and OOD localization.  }
  \label{fig:vis_wo_qd} \vspace{-2mm}
\end{figure} 
\subsection{Ablation Study}
\label{sec:ablation}

{\bf The Effect of the Query-Distribution Loss.}  Without the QD loss, the mean inlier tumor DSC of our framework increases only by a small margin compared to the nnUNet~\cite{isensee2021nnu} baseline (\cref{tab:seg}).  \cref{fig:vis_wo_qd} presents query visualizations to  show benefits from query-level guidance. Most queries redundantly represent the large and heterogeneous region of the background rather than the tumors without the QD loss (\cref{fig:vis_wo_qd} left). With the QD loss, our framework is manipulated to provide fixed resources (queries for tumors) on distinguishing subtle differences of foregrounds for a near-OOD problem (\cref{fig:vis_wo_qd} right). Final results are thus further improved on all metrics by large margins using QD loss (\cref{tab:main,tab:seg}). The results reveal that managing the object queries with QD loss contributes to masking transformers to improve both segmentation and OOD localization/detection performance.

{\bf The Distribution of Queries.} We also perform an in-depth analysis of query distribution, as shown in \cref{tab:query-distribution}. Our method shows robustness to different settings of query distribution. On all settings, our method outperforms the previous leading method, SML~\cite{jung2021standardized}, by a large margin in OOD localization and inlier segmentation. Eventually, we choose the hyper-parameter $(N_1, N_2, N_3)$ as (16, 4, 12).

\begin{table}[t]
\centering
\resizebox{\linewidth}{!}{
\begin{tabular}{l|ccc|c}
\hline
Query Dist. & \multirow{2}*{AUROC$\uparrow$} & \multirow{2}*{AUPR$\uparrow$} & \multirow{2}*{FPR$\downarrow$} & \multirow{2}*{$\mathrm{DSC}_{\mathrm{inlier}}\uparrow$} \\
($N_{1},N_{2},N_{3}$)& & & & \\\hline
SML~\cite{jung2021standardized} &56.10&30.44&77.81&35.47\\
\hline
    (8, 4, 20)        & 84.44      &    51.32  &42.10     &36.50   \\
    (8, 20, 4)        &   83.73    &49.76      &43.32     &39.79     \\
    \textbf{(16, 4, 12)}        &82.52       &\textbf{55.60}      &    46.90 & \textbf{41.77}\\
    (20, 4, 8)        &85.66      &55.17      & 37.24    &38.19     \\
    (24, 4, 4)        & \textbf{86.41}      & 52.58     &\textbf{33.70}   &  39.43   \\
    \hline
\end{tabular}
}
\caption{Ablation study on the distribution of queries. ($\mathrm{DSC}_{\mathrm{inlier}}$: mean Dice Score of inlier tumors.)}
\label{tab:query-distribution}
\end{table}

\begin{table}[t]
\centering
\begin{tabular}{l|c|ccc}
\hline
          Level                      & Softmax     & AUROC$\uparrow$ & AUPR$\uparrow$ & FPR95$\downarrow$ \\
                                \hline
\multirow{2}{*}{Category} & post &  58.14     & 16.28     &  79.29      \\
                          & pre  &  52.70     & 24.59     &  88.40      \\
                                \hline
\multirow{2}{*}{Query}    & post &  76.88     &  33.98    &  55.82     \\
                          & pre  &     \textbf{ 82.52} &\textbf{55.60} &\textbf{46.19 }  \\
                                \hline
\end{tabular}
\caption{Comparison of category- and query-level anomaly scores. With the same network, the query-level anomaly scores show superiority over the category-level ones for OOD localization. Meanwhile, MaxQuery from the pre-softmax query-level scores outperforms that from post-softmax ones.} 
\label{tab:ablation-score}
\vspace{-4mm}
\end{table}

{\bf Pre-softmax versus Post-softmax for MaxQuery.} 
As shown in \cref{tab:ablation-score}, MaxQuery with pre-softmax score $\mathbf{R}$ exceeds the one with post-softmax $\mathbf{M}$ by 21.62\% in AUPR for OOD localization, which agrees with our explanation in \cref{sec:method_ood}.

{\bf Query-level versus Category-level Anomaly Score.}
The debate of pre-softmax versus post-softmax corresponds to the one of MaxLogit~\cite{hendrycks2016baseline} versus MSP~\cite{hendrycks2019scaling}. Specifically, MSP calculates the post-softmax score in the final category level, while MaxLogit calculates the pre-softmax one. Unlike MSP and MaxLogit, our MaxQuery produces an anomalous score at the query level. For a fair comparison, we apply MSP and MaxLogit based on the Mask transformer we used in our model. As shown in \cref{tab:ablation-score},  MaxQuery (post-softmax) outperforms MSP (category, post-softmax) by 17.70\% and MaxQuery (pre-softmax) exceeds MaxLogit (category, pre-softmax) by 32.01\% in AUPR. This comparison indicates the superiority of our query-level anomaly score over the category-level ones.

\section{Conclusion}
\label{sec:conclusion}
Processing a large collection of medical imaging data with long-tailed distributions has always been challenging. The significant performance improvement of our method on two real-world datasets  validates its effectiveness. This result proves that interpreting segmentation as (query) cluster assignment is valid and effective. Our novel MaxQuery and QD loss are also evidently helpful for inlier segmentation and (near-)OOD detection/localization, performing in practical scenarios. We believe that the proposed method has the good potential to further boost the adoption of medical image segmentation in designing various clinical applications.

\section*{Acknowledgement}
 This work was supported by Alibaba Group through Alibaba Research Intern Program. Bin Dong was partly supported by NSFC 12090022.

{\small
\bibliographystyle{ieee_fullname}
\bibliography{egbib}
}
\newpage
\appendix
\section{Appendix}
\renewcommand\thefigure{A\arabic{figure}}
\renewcommand\thetable{A\arabic{table}}

\subsection{Dataset Details}
We provide the abbreviation and full name for each disease from \textit{Pancreatic Tumors} and \textit{Liver Tumors} in \cref{tab:full_name_pancreas,tab:full_name_liver}, respectively. Meanwhile, we report their incidence count in our datasets.

We determine the data splitting of known (inliers) and unknown classes (outliers) according to the real-world medical scenario and previous clinical studies~\cite{chu2022classification,springer2019multimodality}. For \textit{Pancreatic Tumors}, we assign seven common pancreatic diseases (PDAC, PNET, SPT, IPMN, MCN, CP, and SCN) as inliers, and allocate two peri-pancreatic diseases (AC, DC) and ``other" as outliers. The two peri-pancreatic diseases (AC, DC) are relatively difficult to distinguish from PDACs by radiologists, but clinical studies of pancreatic lesion diagnosis~\cite{springer2019multimodality,chu2022classification} did not include them because they are not inside the pancreas. Thus we regard them as OOD in our model. For \textit{Liver Tumors}, we assign five common liver tumors \cite{yasaka2018deep} (HCC, ICC, metastasis, hemangiomas, and cyst) as inliers, and allocate hepatoblastoma, FNH, and ``other" as outliers, due to their low incidental rate.

Note that ``other" class represents rare neoplasms or tumors in the real-world dataset, which reflects the long-tailed distribution of real-world disease incidence. Since these rare diseases are individually infrequent, it is impossible to collect them completely. Therefore, we address the thorny problem by OOD detection and localization.  
\setcounter{table}{0}
\begin{table}[ht]
    \centering
    \resizebox{\linewidth}{!}{
    \begin{tabular}{l|l|c}
    \hline
    Abbr. & Full name & Count\\
    \hline
     PDAC    & Pancreatic ductal adenocarcinoma & 366\\
      IPMN   & Intraductal papillary mucinous neoplasms & 61\\
      PNET & Pancreatic neuroendocrine tumor & 35\\
      SCN & Serous cystic neoplasms & 46\\
      CP &  Chronic pancreatitis & 43\\
      SPT & Solid pseudopapillary tumor & 32\\
      MCN & Mucinous cystadenoma & 7\\
      \hline
      AC & Ampullary cancer & 46\\
      DC & Bile duct cancer & 12\\
      ``other" & Other rare neoplasms & 13 \\
      \hline
    \end{tabular}
    }
    \caption{Dataset details of real-world \textit{Pancreatic Tumors}. This full-spectrum dataset consists of ten pancreatic diseases, among which we assign the top seven as inlier tumors and the bottom three as outlier tumors, based on the real-world medical scenario and previous clinical studies~\cite{springer2019multimodality,chu2022classification}.}
    \label{tab:full_name_pancreas}
\end{table}
\setcounter{table}{1} 
\begin{table}[ht]
    \centering
    \begin{tabular}{l|l|c}
    \hline
    Abbr. & Full name & Count\\
    \hline
     HCC & Hepatocellular carcinoma & 162\\
     ICC & Intrahepatic cholangiocarcinoma & 51\\
     Meta. & Metastasis & 97\\
     Heman. & Hemangiomas & 75\\
     Cyst & Cyst & 146\\
     \hline 
     Hepato. & Hepatoblastoma & 17\\
     FNH & Focal nodular hyperplasia & 27 \\
     ``other" & Other rare tumors & 60\\
      \hline
    \end{tabular}
    \caption{Dataset details of real-world \textit{Liver Tumors}. This full-spectrum dataset includes seven liver tumors, among which we assign the top five as inlier tumors and the bottom three as outlier tumors, according to the real-world medical scenario and previous clinical studies \cite{yasaka2018deep}. }
    \label{tab:full_name_liver}
\end{table}

\subsection{Qualitative Results on Liver Tumors}
For qualitative analysis on \textit{Liver Tumors}, we present visual examples of anomaly score map for OOD localization in \cref{fig:vis_liver}. This shows that our approach achieves a high anomaly score in the OOD pixels (outlier tumor), while a low anomaly score in the in-distribution pixels (organ), compared with other methods.
\setcounter{figure}{0} 
\begin{figure*}[ht]
  \centering
   \includegraphics[width=0.8\linewidth]{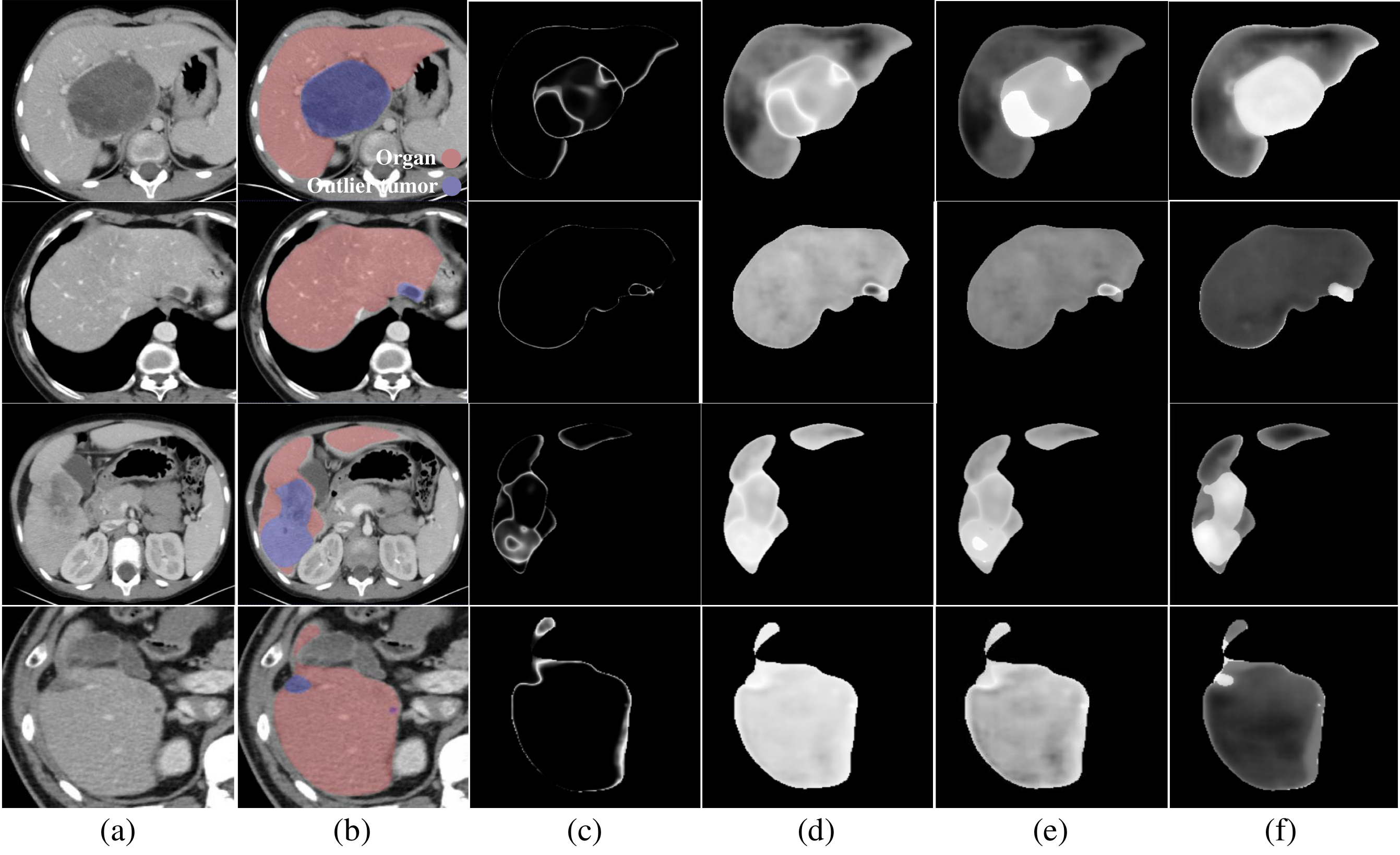}
   \vspace{-0.2cm}
   \caption{Visualization results of anomaly score map for OOD localization on \textit{Liver Tumors}: (a) 2D slices of the CT image, (b) ground truth annotation (red: liver, blue: outlier tumor), (c) MSP~\cite{hendrycks2016baseline}, (d) MaxLogit~\cite{hendrycks2019scaling}, (e) SML~\cite{jung2021standardized} and (f) Ours. The grayscale level indicates the anomaly score. Our method reaches a high anomaly score in the OOD pixels (outlier tumor), while a low anomaly score in the in-distribution pixels (organ).}
   \vspace{-0.2cm}
   \label{fig:vis_liver}
\end{figure*}
\subsection{Baselines for Inlier Segmentation}
\noindent\textbf{Comparison with Other Baselines.} For a fair comparison with our method, we train UNet~\cite{ronneberger2015u_unet0}, UNet++~\cite{zhou2019unet++_unet7}, TransUNet~\cite{chen2021transunet} based on the framework of nnUNet~\cite{isensee2021nnu}. TransUNet adopts transformer modules as pixel encoder, whereas our method uses CNN as the pixel-level backbone and leverages stand-alone transformer modules to interact with it. As presented in \cref{tab:seg_other}, our method shows superiority on inlier segmentation compared with strong baselines, including nnUNet~\cite{isensee2021nnu} and (nn)TransUNet~\cite{chen2021transunet}. This demonstrates that the distinctive architecture of our newly designed mask transformers leads to better performance on real-world medical image segmentation.

We also train Swin UNETR~\cite{tang2022self_swinunetr} using their officially released code and pre-trained model. We find that Swin UNETR~\cite{tang2022self_swinunetr} could not converge to reasonable tumor segmentations on \textit{Pancreatic Tumors}, that might be due to its difficulity in identifying subtle tumor differences without sufficient data samples. Meanwhile, Swin UNETR~\cite{tang2022self_swinunetr} achieves Dice scores of 50.48\% (HCC),   32.62\% (ICC), 36.06\% (Meta.), 71.82\% (Heman.) and 15.30\% (Cyst) on \textit{Liver Tumors}, resulting in the average score of 41.26\%.

\begin{table*}[ht]
\centering
\resizebox{\textwidth}{!}{
\begin{tabular}{l|cccccccc|cccccc}
\hline
           & \multicolumn{8}{c|}{Pancreatic \%}                                         & \multicolumn{6}{c}{Liver \%}                                                      \\ \hline
Methods    & PDAC & IPMN & PNET & SCN & CP & SPT & \multicolumn{1}{c|}{MCN} & Avg. & HCC & ICC & Meta. & Heman. & \multicolumn{1}{c|}{Cyst} & Avg.                  \\ \hline
UNet ~\cite{ronneberger2015u_unet0}      & 63.96 &21.07 & 21.72&30.70 &17.88 &33.96 & \multicolumn{1}{c|}{18.10}&   29.62  &  61.59 & 28.76 &43.77 &65.01 & \multicolumn{1}{c|}{37.39}& 47.30 \\
UNet++~\cite{zhou2019unet++_unet7} &  63.43  &22.85    &  14.52    & 25.09    & 15.02   & 21.36  &\multicolumn{1}{c|}{10.07}&24.62 & 56.51    & 29.13   &  36.88    & 56.74    &\multicolumn{1}{c|}{46.60}&45.17\\
TransUNet~\cite{chen2021transunet} &  64.91    &31.18      &26.78      &38.96     &22.39    &  29.87   &\multicolumn{1}{c|}{30.27}     & 34.91 &52.26    &25.50      &42.31      &    70.90      &\multicolumn{1}{c|}{47.52}     &47.70 \\
nnUNet~\cite{isensee2021nnu}   & 65.65     &  27.60    &32.59      & 36.46    &  23.33  & 31.73  &     \multicolumn{1}{c|}{30.96}    & 35.47     & 57.22    &  28.16   & 52.81      &   77.55     & \multicolumn{1}{c|}{46.49}     &          52.45             \\
\textbf{Ours}       &    67.91     & 46.92     & 32.07     &    42.51 &31.36    &42.67     &      \multicolumn{1}{c|}{28.97}    &   \textbf{41.77}   &  67.61   &  30.78   &  60.40     &  77.07      & \multicolumn{1}{c|}{47.61}     & \textbf{56.69} \\ \hline
\end{tabular}
}
\caption{Inlier segmentation Dice scores (\%) on \textit{val} set of \textit{Pancreatic Tumors} and \textit{Liver Tumors} (all methods report results with final checkpoint). Our method notably outperforms all baselines for the task of inlier tumor segmentation.
}
\label{tab:seg_other} 
\end{table*}

\subsection{Statistical Analysis}
The Wilcoxon signed-rank test shows our method shows significant improvement to the second-best approaches on all metrics with $p<0.01$, as presented in \cref{tab:pvalue}.
\begin{table}[ht]
    \centering
    \resizebox{\linewidth}{!}{
    \begin{tabular}{c|cccc}
    \hline
     $p$   & AUROC    & AUPR   & FPR95    & DSC    \\
         \hline
Pancreas & 4.4$\times 10^{-6}$ & 2.0$\times 10^{-6}$ & 2.7$\times 10^{-7}$ & 2.0$\times 10^{-6}$ \\
Liver    & 2.3$\times 10^{-3}$   & 7.0$\times 10^{-3}$  & 6.7$\times 10^{-3}$   & 2.8$\times 10^{-3}$ \\
\hline
\end{tabular}
    }
    \caption{Results of Wilcoxon signed-rank test versus the second-best approaches on all metrics.}
    \label{tab:pvalue}
\end{table}
\subsection{Hyper-parameter Selection.}
We discuss in detail the key hyper-parameter of our method, i.e., $(N_1, N_2, N_3)$, for controlling the query distribution, in \cref{tab:query-distribution} and \cref{sec:ablation}. Our method shows robustness to different settings of query distribution. And another important hyper-parameter is the number of queries. It should be redundantly larger than the possible/useful classes in the data, which depends heavily on the data and the task. For other hyper-parameters on data augmentation, pre-processing, network architecture, and optimization, we follow the original settings in nnUNet~\cite{isensee2021nnu} and KMax-Deeplab~\cite{yu2022k}.

\end{document}